\RequirePackage{fix-cm}
\documentclass[smallextended]{svjour3}       
\smartqed  
\usepackage{graphicx}
\usepackage{upgreek}
\usepackage{amsmath}
\usepackage{bbm}
\usepackage{times}                     
\usepackage{cite}                      
\usepackage{tabu}                      
\usepackage{booktabs}                  
\usepackage{url}
\usepackage{placeins}
\usepackage{geometry}
\usepackage{lineno}
\begin{document}

\title{A Review, Framework and R toolkit for Exploring, Evaluating, and Comparing Visualizations
}

\titlerunning{A Review, Framework and R toolkit for Exploring, Evaluating, and Comparing Visualizations}        

\author{Stephen France        \and
        Ulas Akkucuk 
}


\institute{Stephen L. France \at
              Assistant Professor of Quantitative Analysis\\
              Mississippi State University\\
              Mississippi State, MS, 39762, USA\\
              Tel.: +1-662.325.1630\\
              Fax: +1-662.325.7012\\
              \email{sfrance@business.msstate.edu}           
           \and
           Ulas Akkucuk \at
            Department of Managment\\
            Bogazici University\\
            Bebek, Istanbul, 34342\\
            Tel.: +90-212-359-54-00\\
            Fax: +90-212-287-78-51\\
            \email{ulas.akkucuk@boun.edu.tr}             
}
\date{Received: date / Accepted: date}

\maketitle

\begin{abstract}
This paper gives a review and synthesis of methods of evaluating dimensionality reduction techniques. Particular attention is paid to rank-order neighborhood evaluation metrics. A framework is created for exploring dimensionality reduction quality through visualization.  An associated toolkit is implemented in R. The toolkit includes scatter plots, heat maps, loess smoothing, and performance lift diagrams.  The overall rationale is to help researchers compare dimensionality reduction techniques and use visual insights to help select and improve techniques. Examples are given for dimensionality reduction of manifolds and for the dimensionality reduction applied to a consumer survey dataset.
\end{abstract}

\section{Introduction}\label{Introduction} 
The problem of dimensionality reduction is core to statistics, machine learning, and visualization. High dimensional data can contain a large amount of noise and importantly for visualization, the human brain can only comprehend three dimensions.  Thus, there is a need to reduce data into an interpretable format by converting high dimensional data into two or three dimensions, which can subsequently be visualized using a two or three dimensional scatterplot.  To meet the need for dimensionality reduction methods, a plethora of algorithms and associated fitting methods have been developed.  A researcher wishing to perform dimensionality reduction for visualization will be presented with a choice of hundreds of algorithms. Which algorithm should be used? This paper describes a visualization framework called QVisVis and associated software tools implemented in R to help choose dimensionality reduction methods, tune these methods, and visually evaluate the quality of dimensionality reduction solutions. The major contributions of these paper are to review and synthesize previous work on evaluating and ``visualizing'' performance metrics, create an overall visualization framework for ``visualizing'' visualization quality, and implement the framework in an R toolkit. 

\subsection{Visualization Design and Evaluation}\label{Design} 
The roots of much of modern data-based visualization come from exploratory data analysis, which was popularized by John Tukey \cite{REF685}, who developed an array of simple tools, such as the box plot, to help summarize, explore, and ultimately gain insight from data. This idea of ``exploration'' is still core to modern visualization. Visualization exploration  \cite{REF1502} can be thought of as a process where a user tunes parameters to transform and explore data.  At each stage of the process, parameters are passed to a visualization transform \cite{REF1503} function, which creates the visualization, which the user then uses to further train parameters, as part of a feedback loop. When implementing data visualization systems, both artistic \cite{REF1504} and data-based engineering considerations come into play.  An overarching consideration, which subsumes both artistic and engineering aspects is that of design \cite{REF1505}. Design considerations are succinctly categorized using the scheme adapted from the Roman era design work of Vitruvius \cite{REF1506}, where design is evaluated on a triad of soundness, attractiveness, and utility.  This design based view can be combined with the previously described process based view in a design activity framework \cite{REF1507}.   Here, the design process is split into activities, which fall into one or more of the categories understand, ideate, make, and deploy. The researcher will try to understand the problem along with its opportunities and constraints and then generate ideas, which are then implemented and finally deployed.

Throughout all of the stages of the visualization process and in particular the ideate and make stages, there is a need to evaluate visualizations.  There are several methods and frameworks for evaluating visualizations, some qualitative and some quantitative \cite{REF1512}, which include case studies, controlled experiments and usability tests \cite{REF1511}.  One qualitative framework is the value based evaluation framework \cite{REF1508}. Here, a visualization's value is expressed as a function of how a user can find insights in the minimal amount of \textbf{time}, discover \textbf{insights}, uncover the \textbf{essence} or sense of the data, and generate \textbf{confidence} in the data and data domain. The ability to generate insight is particularly important, but as ``insight'' is a qualitative, many-faceted concept, it can be difficult to measure, requiring deep, open ended exploration, with an emphasis on the problem domain \cite{REF1509}.  Another approach to visualization evaluation is to look at the validity of the different abstracted steps of visualization process \cite{REF1510}, ranging from the high level problem characterization, through data/operations, encoding/interaction, and finally algorithm design. At each stage, threats to validity may occur and these must be dealt with.  From an evaluation perspective, there may not be one single optimal visualization and visualization evaluation will involve a trade-off between different visualization criteria.  For example, there is often a trade-off between optimizing local visualization features vs. the overall global consistency of a visualization \cite{REF1513}.

\subsection{Evaluation for Dimensionality Reduction}\label{Evaluation} 
This paper deals with a specific type of visualization evaluation. When data are transformed from a higher number of dimensions to an embedding in a lower number of dimensions, the projection is not always perfect, due to reduced degrees of freedom in the destination embedding.  For example, an item may be the nearest neighbor of many different points in high dimensional space, but only two points in one dimensional space \cite{REF1514,REF1517}. This issue has been apparent from the early days of visualization.  It is impossible to perfectly represent a three dimensional globe on a two dimensional map and keep a perfect representation of distances. Thus, throughout history, a range of different map projections have been created \cite{REF398}, each with some distortion of actual distances.  Different map projections have different types of distortion. For example, the Mercator projection is conformal, i.e., local angles and shapes are preserved, but distances at higher latitudes are distorted and seem much greater than actuality. An equirectangular projection, preserves distances along meridians, but is not conformal.  The Mercator projection is widely used in navigation, where the conformal property is important.  The equirectangular projection, because of a lack of conformality, cannot be used in navigation, but is often used in raster visualization, due to its relatively even global spacing. The decision to use a specific map visualization is essentially a trade-off, with different types of distortion being more or less problematic for different applications.  A projection is chosen in order that the map distortion has minimal effect in the problem domain.  The level of distortion at different areas of the mapping can be visualized using Tissot's Indicatrix \cite{REF1516}, which plots ellipses at regular intervals on the map to show distortion relative to a unit sphere.

In the context of dimensionality reduction, the final output (usually in two or three dimensions) is often used for visualization.  However, visualization can be utilized to help understand and analyze the dimensionality reduction process.  For example, the DimStiller system \cite{REF1618} utilizes scree plots, inter-variable correlations, and scatterplot matrices of derived solutions colored by class to help inform and optimize the dimensionality reduction process. Visualization can be used at all stages of the visualization process including the dimensionality reduction, visualization, and user analysis stages \cite{REF1620}, with simple visualizations such as scatterplots and parallel coordinates approaches being particularly useful\cite{REF1619}.

The QVisVis framework described in this paper includes several techniques to help evaluate the quality of visualizations created using dimensionality reduction techniques. These tools utilize quantitative measures of item neighborhood recovery, building on the ideas of nearest neighborhood recovery \cite{REF1514,REF1517}.  In a similar fashion to Tissot's Indicatrix, visual representations help quantify and visualize the distortions from perfection in the lower dimensional data visualization. A range of visual tools are included and together they are utilized to give a broad view of visualization performance, help tune parameters for dimensionality reduction algorithms, and help users understand the trade-offs involved in visualizing data, particularly with respect to the trade-off between local and global data quality. The methods can be used at the algorithm/design stage of the visualization process, or at a higher level in the abstraction stage, to help comprehend the relationship between the source data attributes and visualization performance.  The remainder of this paper follows the following structure.  First, a brief overview of the dimensionality reduction problem is given, along with a summary of several classic dimensionality reduction algorithms. Then, rank-order methods of calculating solution agreement metrics are reviewed. This is followed by a description of the different visualizations included in the framework. Visualization examples are given for classic dimensionality reduction methods performed in 3 dimensional manifold.  A real world data example is given, where consumer preference survey data with 135 dimensions is visualized in two dimensions.  Here, the t-SNE (stochastic neighborhood embedding) method is compared with two classic dimensionality reduction methods for multiple parameter settings.

\section{Dimensionality Reduction}\label{DimReduct} 

This section gives a brief overview of dimensionality reduction. A more detailed exposition is available in a detailed comparative review \cite{REF1518}.

Let  ${\bf{X}} = (x_{il})_{\{n\times m \}}$ be a source matrix of data for \textit{n} items given in \textit{m} dimensions.  Here $x_{il}$ is the value of item \textit{i} on dimension \textit{j}.  The data can be transformed into a proximity matrix ${\bf{P}} = (p_{ij})_{\{n\times n \}}$, where $d_{ij}$ is the proximity between items \textit{i} and \textit{j}.  This measure can either be a similarity $s_{ij}$ or a distance $d_{ij}$, where $s_{ij}\geq s_{kl}$ implies that $d_{ij}\leq d_{kl}$. A range of similarity measures and dissimilarity measures can be employed.  The exact measure will depend on the data set characteristics, such as size, data type, dimensional and sparsely. A common measure of similarity is the correlation, which is given in  (\ref{eq:Correlation}).  A common measure of distance is the Euclidean distance, which is given in (\ref{eq:Euclidean}) and is a special case for $p=2$ of the $L^p$ or Minkowski norms used to define vector spaces \cite{REF1521}. 
\begin{equation}
\label{eq:Correlation}
s_{ij}=\frac{\sum_{l=1}^{m}\left(x_{il}-\bar{\bf{x}}_i\right)\left(x_{jl}-\bar{\bf{x}}_j\right)}{\sqrt{\sum_{l=1}^{m}\left(x_{il}-\bar{\bf{x}}_i\right)^2}\sqrt{\sum_{l=1}^{m}\left(x_{jl}-\bar{\bf{x}}_j\right)^2}}
\end{equation}
Here, $\bar{{\bf{x}}}_i$ is the mean value for vector ${\bf{x}}_i$.
\begin{equation}
\label{eq:Euclidean}
d_{ij}=\sqrt{\sum_{l=1}^{m}\left(x_{il}-x_{jl}\right)^2}
\end{equation}
Parameterized distance metrics can be optimized or ``learned'' for optimal performance on a given dataset \cite{REF1520}. These learned metrics often perform better than standard metrics.

\subsection{PCA and Classical MDS}\label{PCA} 
One of the classic, most commonly used dimensionality reduction methods is PCA (principal component analysis) \cite{REF304,REF1498}. PCA can take noisy high dimensional data and convert these data into lower dimensional data that account for the maximum possible variance and have mutually uncorrelated dimensions.  There are two major variants of PCA. The first utilizes the covariance matrix and the second the correlation matrix.  A correlation matrix is used if one wishes to normalize the variance across items and a covariance matrix is used if one wishes to account for this variance in the analysis.  Define a similarity matrix \textbf{S}, where for a correlation matrix, $s_{ij}$ is given in (\ref{eq:Correlation}) and for a covariance matrix, $s_{ij}$ is the numerator of  (\ref{eq:Correlation}).  This matrix is sometimes referred to as the scalar products matrix.  An eigendecomposition ${\bf{S}}={\bf{Q\Lambda Q'}}$ is then performed.  Here, $\bf{\Lambda}$ contains a set of diagonalized eigenvalues, which give the variance explained by each derived dimension. The lower dimensional solution for \textit{k} dimensions is ${\bf{V}}={\bf{Q\Lambda}}_k^{1/2}$.  The values of $\lambda_i$ where \textit{i} is greater than the dimensionality \textit{k} in ${\bf{\Lambda}}_k$ are set to be equal to 0.

In some applications, the source data may be directly gathered similarity or dissimilarity data, which can occur when users directly rate the similarity or dissimilarity of items or where similarity is derived from some measure of co-occurrence, such as shopping items being purchased together. Consider some dissimilarity matrix ${\bf{\Delta}}_{\{n\times n \}}$ where $\delta_{ij}$ is the directly derived dissimilarity between items \textit{i} and \textit{j}. By ensuring that the dissimilarities meet the basic distance axioms and triangle inequality \cite{REF1522}, ${\bf{\Delta}}$ can be considered to be a distance matrix ${\bf{D}}$. The method of Classical MDS \cite{REF53,REF54} converts a dissimilarity/distance matrix into a scalar products matrix, using the double centering formula given in  (\ref{eq:ScalarProduct}).
\begin{equation}
\label{eq:ScalarProduct}
{\bf{S}}={\bf{D}}-{\bf{\bar{D}}}_{i\cdot}-{\bf{\bar{D}}'}_{i\cdot}+{\bf{\bar{\bar{D}}}}_{\cdot\cdot}
\end{equation}
Here, ${\bf{\bar{D}}}_{i\cdot}$ is a matrix of row (or column means) of \textbf{D} and ${\bf{\bar{\bar{D}}}}_{\cdot\cdot}$ is a matrix of overall distance means. The resulting scalar products matrix can be analyzed using the same method as the scalar products matrix in PCA.

\subsection{Distance based MDS}\label{DistMDS} 
An alternative to classical MDS is distance based MDS \cite{REF50,REF51}, which is often used in visualization applications \cite{REF282,REF886}. Here data are reduced to a predefined lower dimensional solution using a loss function called ``Stress'' that minimizes the differences between the source distances/dissimilarities and the distances in the lower dimensional solution.  The Stress function is given in (\ref{eq:Stress}).
\begin{equation}
\label{eq:Stress}
Stress = \sqrt {\frac{{\sum\limits_i {\sum\limits_{j \ne i} {{{{w_{ij}}\left( {{d_{ij}}
- {{\hat d}_{ij}}} \right)}^2}} } }}{{\sum\limits_i {\sum\limits_{j \ne i}
{w_{ij}}{d_{ij}^2} } }}}
\end{equation}
Here, each ${\hat d}_{ij}$ is a transformed input dissimilarity  ${\hat d}_{ij}=f\left(\delta_{ij}\right)$.  The transform function is dependent on the level of the data.  For ordinal scale data, a monotone transformation can be used \cite{REF51} and for continuous data a simple shift/scaling transformation can be used.  A wide range of transformations, including spline transformations are available \cite{REF277}, along with a range of variants of the Stress function \cite{REF1523}, for different data and usage scenarios. The value of $d_{ij}$ is the Euclidean distance between points \textit{i} and \textit{j} in the lower dimensional derived solution. The Stress function can be optimized using a simple gradient descent or majorization optimization method \cite{REF120}. The weights $w_{ij}$ by default are set to 1 for present data and 0 for missing data.  However, weights can be altered for domain specific importance or to alter the trade-off between local and global recovery.  For example, there are variants of local multidimensional scaling \cite{REF352,REF464}, where weights $w_{ij}$ are higher for local distances and can be set to 0 if distances are greater than a certain threshold. 

\subsection{Local Linear Embedding}\label{LLE} 
The methods discussed thus far are global methods.  They utilize global fitting procedures that attempt to accurately reconstruct both smaller local distances and larger global distances. There are several methods that prioritize local fitting. One method is LLE (Local Linear Embedding) \cite{REF402}. Each point in a configuration ${\bf{x}}_{i}$ can be represented using a weighted linear combination of nearby points. A minimization function to set the weights is given in (\ref{eq:LLEHigher}).
\begin{equation}
\label{eq:LLEHigher}
\min \sum_{i=1}^{n}\left\Vert{\bf{x}}_{i}-\sum_{k=1}^{P}w_{ij}{\bf{x}}_{j}\right\Vert
\end{equation}
Here, the values of $w_{ij}$ are set to minimize the difference between each ${\bf{x}}_{i}$ and its linear combination of nearby points using constrained least squares optimization.  The size of the neighborhood for each item is an input parameter to the LLE procedure. The fitted weights are then used to create a derived lower dimensional solution using (\ref{eq:LLELower}).
\begin{equation}
\label{eq:LLELower}
\min \sum_{i=1}^{n}\left\Vert{\bf{v}}_{i}-\sum_{k=1}^{P}w_{ij}{\bf{v}}_{j}\right\Vert
\end{equation}
Here, ${\bf{v}}_{i}$ is the lower dimensional output vector for item \textit{i}.  The total system of equations defined in (\ref{eq:LLELower}) can be solved as a sparse eigenvector problem. 

In summary, a range of dimensionality reduction techniques have been described in this section.  These techniques were chosen both for popularity, the quality of solution, available implementation software, and parameterizion options that allow the local/global recovery trade-off to be altered, a concept that is at the core of the visualization framework described in this paper.

\subsection{Additional Techniques}\label{Additional} 
The four basic techniques described in the previous section illustrate several aspects of the dimensionality reduction problem.  With dimensionality reduction techniques, there is a trade-off between local and global solution recovery. Some methods, such as LLE, focus on local recovery, while others focus on global reconstruction \cite{REF514}.  The optimal technique very much depends on the application.  In some instances, local neighborhood accuracy is important, for example, when dealing with nonlinear manifolds \cite{REF1524} or mapping applications where the user of a visualization wishes to ask questions such as ``which is the nearest neighbor of item \textit{i}?''.  In other cases, for example when creating a ``bigger picture'' global map, the overall global scale and accuracy is important.  These trade-offs will be explored in the subsequently described QVisVis framework. Another aspect of any dimensionality reduction technique is the fitting or optimization algorithm.  The methods described thus far utilize some optimization procedure, either using a nonlinear optimization algorithm or an eigenvector decomposition.  In the case of LLE, a parameter (neighborhood size) could be utilized to tune the procedure, creating a trade-off between different aspects of solution quality. This idea of parameter tuning is explored further when describing the overall visualization framework.

In addition to the basic techniques above, several other techniques were implemented to demonstrate the QVisVis framework.  These were chosen both for popularity, the ability to cope with a wide range of datasets, and strong conceptual relationships with the basic techniques. Isomap \cite{REF173} is a variant of classical MDS designed for nonlinear data. Isomap creates a network, where items are joined if they are within a certain neighborhood of one another. The neighborhood is defined by either a distance $\epsilon$ or by the number of nearest neighbors \textit{k}. Shortest path network distances between items are then calculated using an algorithm such as Dijkstra's algorithm \cite{REF525}. The rationale behind the technique is that by taking network distances rather than straight Euclidean distances, it is easier to extract nonlinear surfaces embedded in higher dimensional space. The size of the neighborhood is a trade-off. Taking neighborhoods of size $k=n-1$ that include every other point in the configuration results in Euclidean distances and a global fitting method. Conversely, taking small neighborhood sizes could result in very poor global recovery and if too small, disconnected graphs that cannot be processed.  The Laplacian eigenmap method \cite{REF513} has the same initial step of creating a neighborhood graph and the same tuning parameters as the Isomap method. A matrix of weights  ${\bf{W}} = (w_{ij})_{\{n\times n \}}$ are then defined either simply with $w_{ij}=1$ if items are connected and $w_{ij}=0$ if items are not connected. For a more complex scheme, the weights for connected items are given as (\ref{eq:LapWeight}) and are parameterized by \textit{t}, the kernel smoothing parameter. 
\begin{equation}
\label{eq:LapWeight}
w_{ij}=\frac{-\left\Vert{\bf{x}}_{i}-{\bf{x}}_{j}\right\Vert}{t}
\end{equation}
If ${\bf{\Delta}} = (\delta_{ij})_{\{n\times n \}}$ is a diagonal matrix where $\delta_{ii}=\sum_{j=1}^n {d_{ij}}$ and the graph Laplacian is defined as ${\bf{L}}={\bf{\Delta}}-{\bf{W}}$,  then the process of finding a lower dimensional solution can be summarized in the generalized eigenvector equation given in (\ref{eq:Laplacian}). 
\begin{equation}
\label{eq:Laplacian}
{\bf{L}}{\bf{f}}=\boldmath{\lambda}{\bf{\Delta}}{\bf{f}}
\end{equation}
Given possible eigenvalues from $\lambda_0,\lambda_1,\cdots,\lambda_{n-1}$ and a required \textit{k} dimensional solution, the eigenvectors $\left[{\bf{f}}_1\cdots{\bf{f}}_k\right]$ form the lower dimensional solution.  There are several similar methods and variants for Laplacian eigenmaps.  For example, the diffusion maps method \cite{REF1532} utilizes a similar formulation to the Laplacian Eigenmaps method and uses a similar eigenvector decomposition, but is based on a slightly different kernel Laplacian distance framework.

\section{Solution Agreement}\label{SolutionAgree} 
Each of the dimensionality reduction methods described in the previous section implements some goodness of fit criterion.  These criteria are measured on different scales and utilize different fitting methodologies, so cannot be compared directly. Thus, some method independent measure of solution fit is required.  Rank order methods of solution agreement that measure item neighborhood recovery provide such a criteria independent measure of solution quality.  Measures of solution agreement based on the Rand index \cite{REF435} and the Hubert-Arabie adjusted Rand index \cite{REF169,REF863} for clustering agreement have been developed to measure dimensionality reduction solution quality \cite{REF160,REF161,REF406,REF352}. In addition, continuous measures have been used to measure changes in relative distances, such as point compression and stretching \cite{REF1616}.

\subsection{The Agreement Metric}\label{AgreeMetric} 
Given some solution configuration ${\bf{A}} = (a_{ij})_{\{n\times m \}}$ and a distance metric function $f({\bf{A}})={\bf{D}}_A$, a ranking matrix ${\bf{R}}_A=(r_{Aij})_{\{n\times n \}}$ can be derived from ${\bf{D}}_A$, where for row \textit{i}, $r_{Aij}$ is the ascending ranking of $d_{ij}$ on row \textit{i}, excluding $d_{ii}=0$. The neighborhood matrix  ${\bf{N}}_A = (n_{Aij})_{\{n\times m \}}$ gives the indexes for the rankings, so that for row \textit{i} and column \textit{j}, $n_{Aij}=l$, where $r_{Ail}=j$. Consider two neighborhood configurations {\bf{A}} and {\bf{B}} with neighborhood matrices defined as ${\bf{N}}_A$ and ${\bf{N}}_B$. In a dimensionality reduction context, {\bf{A}} is the higher dimensional solution and {\bf{B}} is the derived solution. Let $a_{ik}$ be the number of elements shared by the first \textit{k} columns of ${\bf{N}}_A$ and ${\bf{N}}_B$. The agreement for a neighborhood of size \textit{k} is be defined as (\ref{eq:Agreement}).
\begin{equation}
\label{eq:Agreement}
AR_k = \frac{1}{{kn}}\sum\limits_{i = 1}^n a_{ik}
\end{equation} 
If $k=1$ then $AR_k$ is a local nearest neighbor measure of recovery.  If $k=n-1$ then $AR_k$ is a global measure of neighborhood recovery across all neighborhood sizes.  The measure can be adjusted to account for random agreement either by sampling randomly from an empirical distribution  \cite{REF160,REF161} or by calculating expected agreement using a hypergeometric distribution \cite{REF406,REF352,REF57} as $E(AR_k)=\frac{k}{n-1}$.  The subsequent adjusted agreement is given as (\ref{eq:ExAgreement}).
\begin{equation}
\label{eq:ExAgreement}
AR^*_k = \frac{1}{kn}\sum\limits_{i = 1}^n\left[a_{ik}-\frac{k^2}{n-1}\right]
\end{equation}
The agreement can be summed across all possible values of  $k=1,\cdots,n-1$. The value of $AR^*_k$ over \textit{k} is bounded by 1.  A measure of quality, given in \cite{REF57}, is the proportion of the area across $k=1,\cdots,n-1$ between the random value $E(AR_k)=\frac{k}{n-1}$ and 1 that is below the $AR^*_k$ function\footnote{This value is usually in $\left[0,1\right]$.  However, it is possible that if there is less agreement than random, this value is less than 0.}. This value is denoted as $\psi$ and is given in (\ref{eq:psi}). It is conceptually similar to the receiver operating character (ROC) curve \cite{REF1628} measures utilized for analyzing classification performance \cite{REF1629,REF1630}. 

\begin{equation}
\label{eq:psi}
\psi  = \frac{\sum\limits_{k = 1}^{n-1}AR^*_k}{\sum\limits_{k = 1}^{n-1}\frac{n-k-1}{n-1}}
\end{equation}
A function weighted variant of $\psi$ is given in (\ref{eq:psifw}). 
\begin{equation}
\label{eq:psifw}
\psi_{f\left(k\right)}  = \frac{\sum\limits_{k = 1}^{n-1}f\left(k\right)AR^*_k}{\sum\limits_{k = 1}^{n-1}f\left(k\right)\frac{n-k-1}{n-1}}
\end{equation}
A simple indicator function can be used to find $\psi_{f\left(k\right)}$ for some subset of $k=1\cdots n-1$.  More complex functions can be utilized.  Consider a situation where nearest neighbors are important, for example, finding competing brands in a brand mapping exercise.  Here, local recovery is important, but some medium range recovery is required. For the first third of the remaining $n-1$ items, a full weighting is utilized, which then linearly decreases to 0 over the second third of the items. The resulting weighting function $f\left(k\right)$ is given in (\ref{eq:fkexample}).
\begin{equation}
\label{eq:fkexample}
f\left(k\right)  = 
        \begin{cases}
            1 & \text{if $1\leq k< \left \lfloor{\frac{n-1}{3}}\right \rfloor$} \\
            1-\frac{k-\left(n/3\right)}{\left(n/3\right)}& \text{if $ \left \lfloor{\frac{n-1}{3}}\right \rfloor\leq k< \left \lfloor{\frac{2\left(n-1\right)}{3}}\right \rfloor$} \\
            0 & \text{otherwise}
        \end{cases} 
\end{equation}

\subsection{Extensions}\label{Extensions} 
The agreement metric for a single value of \textit{k} is general for any two configurations ${\bf{A}}$ and ${\bf{B}}$. In practical usage, it can be thought of as analogous to a correlation coefficient.  In fact, a partial agreement coefficient can be implemented in a similar manner to a partial correlation coefficient \cite{REF887}, where a third configuration ${\bf{Z}}$ is believed to influence the relationship between ${\bf{A}}$ and ${\bf{B}}$. Here, the agreement rates between ${\bf{A}}$ and ${\bf{Z}}$ and ${\bf{B}}$ and ${\bf{Z}}$ are removed from the relationship.
The agreement measure is symmetric in that $AK_k$ is only incremented if an item is in the neighborhoods of both ${\bf{A}}$ and ${\bf{B}}$.  While ${\bf{A}}$ may be a higher dimensional solution and ${\bf{B}}$ a lower dimensional solution, the order does not matter.  There are variants of the agreement metric where solution order does matter and where asymmetric solution quality is measured by counting items in the higher dimensional neighborhoods, but not in the lower dimensional neighborhoods and vice versa \cite{REF1526,REF434,REF1527}. Both symmetric and asymmetric agreement metrics are brought together elegantly using the idea of a co-ranking matrix \cite{REF632,REF1600,REF1526}.

Consider a higher dimensional solution ${\bf{A}}$ and an output embedding ${\bf{B}}$, with neighborhood matrices ${\bf{N}}_A$ and ${\bf{N}}_B$, with a co-ranking matrix defined as $\mathbf{\Omega}=(\omega_{ij})_{\{n\times n \}}$.  The value of $\omega_{ij}$ is the number of items that are in column \textit{i} in ${\bf{N}}_A$ and in column \textit{j} in ${\bf{N}}_B$, i.e., $\omega_{ij}=\sum\limits_{l=1}^{n}\sum\limits_{p=1}^{n-1}\mathbbm{1}\left(\left(n_{Ali}=p\right)\wedge\left(n_{Blj}=p \right)\right)$.
For a given item \textit{i}, if another item \textit{j} is a nearer neighbor to \textit{i} in the output solution than in the output solution then this is known as an ``intrusion'' into the solution. For the converse, this is an ``extrusion''.  For a given value of \textit{k}, the intrusion or extrusion is defined as soft if $a_{ij}$ and $b_{ij}$ are on the same side of \textit{k} and hard if \textit{k} splits the values. A summary is given below.
\begin{enumerate}
\item{Hard Intrusion:  $b_{ij} \le k < a_{ij}$}
\item{Soft Intrusion:   $b_{ij} <  a_{ij} \le k$}
\item{Hard Extrusion:  $a_{ij} \le k < b_{ij}$}
\item{Soft Extrusion:  $a_{ij} < b_{ij} \le k$}
\end{enumerate}

To conclude, a range of metrics is available for calculating neighborhood or rank order agreement between solution configurations. The basic agreement metric is symmetric, but non-symmetric metrics can be utilized.  Both symmetric and non-symmetric metrics are summarized by a co-ranking matrix.  There are several dimensions of agreement.  Agreement can range over the size of neighborhood \textit{k} or by the position in either the input or the output visualization. In fact, item specific agreement can be calculated by only including the terms for a specific item.

\section{A Visualization Framework for Agreement}\label{Framework} 

Much of the previous work on agreement metrics has included simple visualizations to view and explore the level of agreement. The simplest type of visualization is a line plot, where \textit{k} is plotted on the abscissa and the level of agreement is plotted on the ordinate. This graph can be shaded to give an area plot of the proportion of the available ``lift'' gained over random agreement \cite{REF57}. Multiple dimensionality reduction techniques can be compared against one another, along with the expected level of agreement. Another method is to plot the lower dimensional embedding and shade by the level of neighborhood recovery for each of the plotted items \cite{REF1529,REF1528}. This type of neighborhood visualization can help identify regions of the visualization where neighbor recovery is poor. Furthermore, to examine relative performance over time, a heatmap can be produced, showing the level of agreement either over \textit{k} for each item or across the co-ranking matrix \cite{REF631,REF1528}.

\subsection{Basic Taxonomy}\label{Taxonomy} 
This section describes a simple taxonomy for dimensionality reduction visualization.  Beyond the models for visualization design and evaluation described previously, several different types of design taxonomy have been described in the visualization literature, including a framework that classifies algorithms based on design considerations and discrete and continuous display attributes \cite{REF1530} and a framework that cross-classifies data features with visualization design features \cite{REF1530}. 

The purpose of the taxonomy is to summarize the previously discussed aspects of dimensionality reduction visualization into an abstract framework.  The elements of the framework define both the type of display, the level of information contained in the display, and the type of insight to be gained from the data.  Combining the visualization display type with the insight type provides the template for the visualization. We draw from both of the above frameworks, though for dimensionality reduction applications, the data types are relatively fixed.  The overall summary of the aspects of algorithm performance to be visualized are given in Table \ref{tb:DesFramework}.  

\begingroup
\setlength{\thinmuskip}{0mu}
\begin{table}[!htb]
\centering
\caption{Summary of Design Framework}
\label{tb:DesFramework}
\begin{tabular}{lp{3cm}p{11cm}} \hline
Name&Values&Desc.\\ \hline
Config&${\bf{A}}$,${\bf{B}}$,both&Whether configuration ${\bf{A}}$, ${\bf{B}}$, or both is/are plotted. The chosen configurations should have 2 or 3 dimensions\\
Comp&Simple or Compare&If simple then an absolute measure of agreement is given. If compare, multiple methods (and ${\bf{B}}$'s) are compared.\\
Eval&Hard, Soft, Both&Whether or not hard, soft, or both intrusions/extrusions are included.\\
Adjust&No, Yes&Whether or not the adjusted variant of the agreement metric is used.\\
Range(k)&$Sub\left(k\right)\subseteq 1\cdots n-1$&Helps define the local to global trade-off. The subset of k-values plotted.\\
Aggr&All,Item&Whether or not the aggregate agreement is plotted for all points or for each item individually.\\ 
Param&Single,Multiple&Whether or not a single parameter set is used or multiple parameter sets are used.\\ \hline
\end{tabular}
\end{table}
\endgroup

The actual graphics plotting techniques are kept separate from the framework, which is at a slightly higher level of abstraction.  Different graphics methods are able to deal with different levels of the framework.  For example, for a scatterplot, it would not be visually useful to aggregate agreement across all points.  

\section{Visualization Examples}\label{Examples} 
To demonstrate the use of the framework, six different manifolds were created, each with 1000 points in 3 dimensions. These were two spheres, one with points mapped at regular intervals and one with points generated randomly, a Swiss roll, and three toruses, one with randomly spaced points, one with a large ring diameter with regularly spaced points, and one with a small ring diameter with regularly spaced points. Each of the three dimensional manifolds was run using several different dimensionality algorithms in R. The algorithms were chosen to give a variance in terms of solution quality and different levels of trade-off between local and global recovery.  The algorithms are listed below.
\begin{enumerate}
\item{{\bf{PCA}}:  The basic PCA algorithm was implemented with the Princomp package in R.}
\item{{\bf{Smacof Distance MDS}}:   The basic distance based MDS procedure was implemented with the smacof package in R \cite{REF452}}
\item{{\bf{Local MDS}}:Implements Smacof distance MDS as above, but with weights $w_{ij}=1$ for the lowest 10\% of distances and $w_{ij}=0$ for all other distances.}
\item{{\bf{LLE}}:Local linear embedding, implemented using the dimRed package \cite{REF1434} in R.}
\item{{\bf{Isomap}}:Implemented using the dimRed package in R.}
\item{{\bf{Laplacian Eigenmaps}}: Implemented using the dimRed package in R.}
\item{{\bf{Diffusion Maps}}: Implemented using the DiffusionMaps package in R.}
\item{{\bf{kPCA}}: Kernel PCA \cite{REF441} is a variant of PCA, where the scalar product distances are transformed using a distance kernel \cite{REF441} in order to allow PCA  to deal with nonlinear data. Implemented with the kernlab package \cite{REF1533} in R.}
\end{enumerate}

\subsection{Scatter Plots}\label{ScatterPlots} 
Scatterplots are a core element of exploratory data analysis.  Scatterplots enable users to see and interpret patterns in the data. They are used as both an initial step to evaluate data assumptions and correlations before running inferential statistical models and also to help analyze patterns for model parameters and residuals after an analysis has been run.  

Scatterplots are commonly used to analyze visualizations with respect to data clusters. For example, in \cite{REF1605}, colored scatterplots are utilized to show a range of different data class and neighborhood validation metrics and in \cite{REF1599}, scatterplots are colored by neighborhood recovery and by the network centrality of plotted points.  In \cite{REF1617}, PCA-biplots (scatterplots with vectors showing the original dimensions) are used to analyze the projection of clusters in three dimensions into two dimensions.  Beyond the methods described in Section \ref{Evaluation}, several of the papers describing methods of solution agreement include plots of solution points colored by local values of the agreement metrics \cite{REF631,REF1529,REF1528}.

The QVisVis framework described in this paper allows for a range of scatterplots to be created.  Both 2D and 3D scatterplots can be created. The scatterplots are colored based on the level of agreement.  Three types of agreement coloring are available. The first is absolute coloring, where the level of agreement goes from 0 to a maximal level of agreement. The second is relative to random agreement, where the expected random agreement is subtracted from the level of agreement. The third is comparative coloring, where the difference in agreement between two configurations is taken.

\begin{figure}[htb]
 \centering 
 \includegraphics[width=0.9\columnwidth]{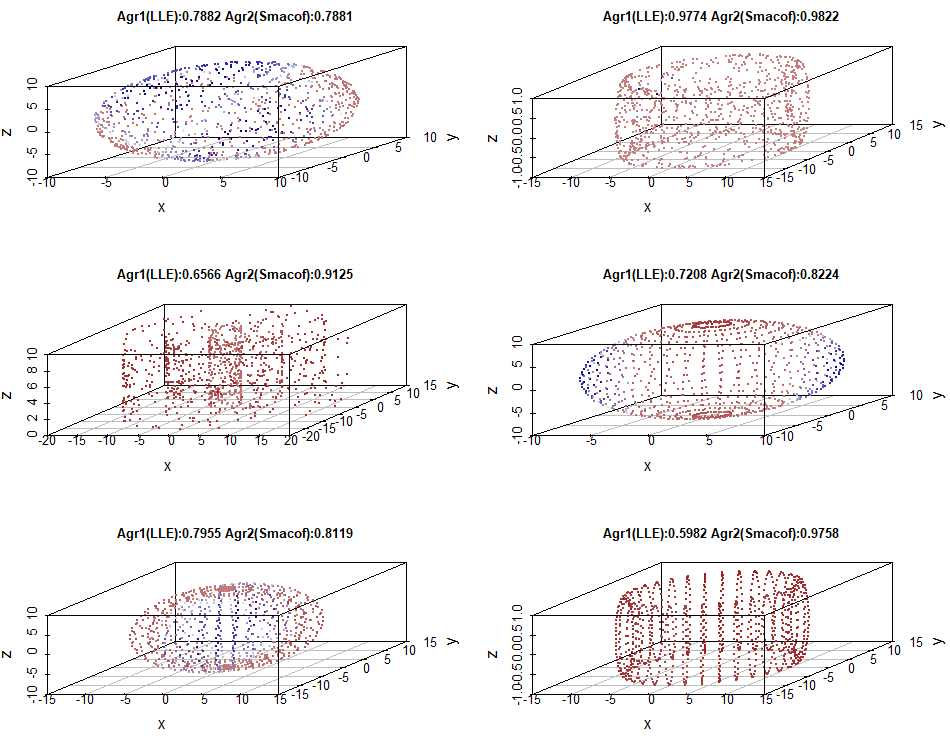}
 \caption{3D LLE vs. Smacof across all values of \textit{k}.}
 \label{fig:3DComp1}
\end{figure}

\begin{figure}[htb]
 \centering 
 \includegraphics[width=0.9\columnwidth]{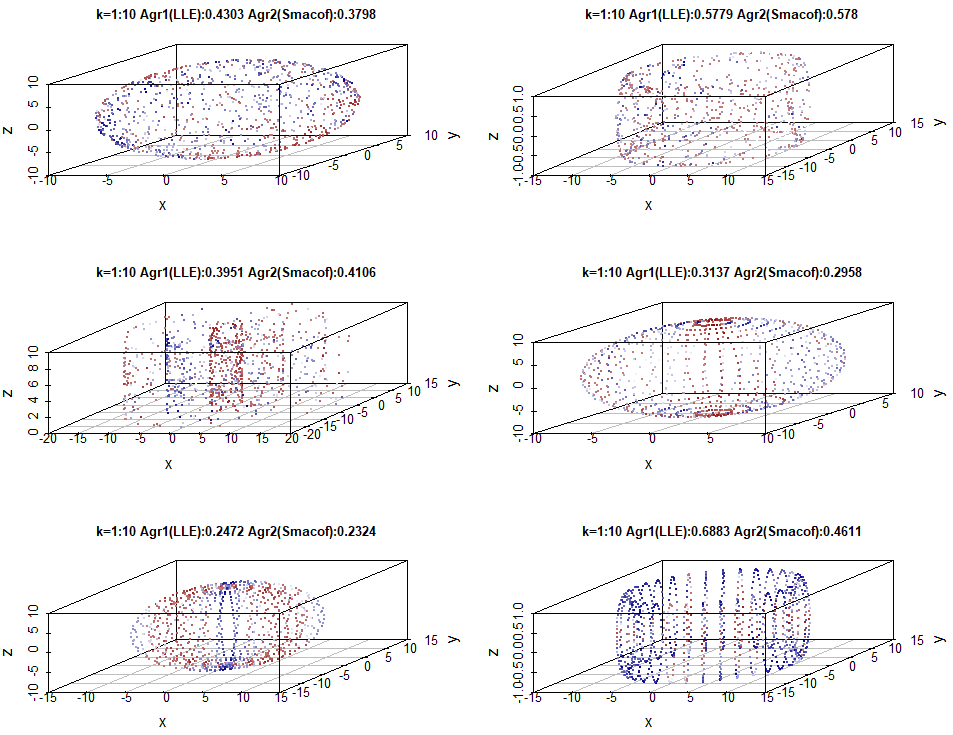}
 \caption{3D LLE vs. Smacof for $k=1\cdots10$.}
 \label{fig:3DComp2}
\end{figure}

The first example shows the difference between a local and global method. Figure \ref{fig:3DComp1} shows the comparative agreement between LLE and Smacof for each manifold across all \textit{k} on the original higher dimensional configuration.  If the average agreement for an item is greater for the first technique (LLE) than the second technique (Smacof) then the point is shaded blue.  If the opposite occurs then the point is shaded red. Greater differences give darker colors.
One can see that the average agreement for Smacof is greater than LLE on all of the manifolds except for the circle with random points.  For this circle, some areas have better recovery with Smacof and others have better recovery with LLE. The third torus, with the small diameter and the regularly spaced rings has particularly bad recovery with LLE, which is probably because the local recovery algorithm does not manage to properly recreate distances between the rings.   The second torus, the large diameter torus is interesting, as the interior points have better performance with LLE and the exterior points have better performance with Smacof.
Figure \ref{fig:3DComp2} shows a similar diagram, but is restricted to local neighborhoods for $k=1\cdots10$.
The pattern here is a little different.  The overall agreement values are much lower. This is to be expected, as to get perfect local neighborhood recovery of 10 points when there are 1000 points in each configuration is very difficult. Also, the relative performance of LLE vs. Smacof is much stronger than for the all \textit{k} example, which again is to be expected, as LLE is a local method and Smacof is a global method. For most of the configurations, the performance between LLE and Smacof is very similar.  The exception is the small diameter torus, which in this instance, has much better recovery for LLE than for Smacof.  This is probably because LLE can recreate the local within ring distances, but for Smacof, there may be some overlap/interference between the rings.

\begin{figure}[htb]
 \centering 
 \includegraphics[width=0.9\columnwidth]{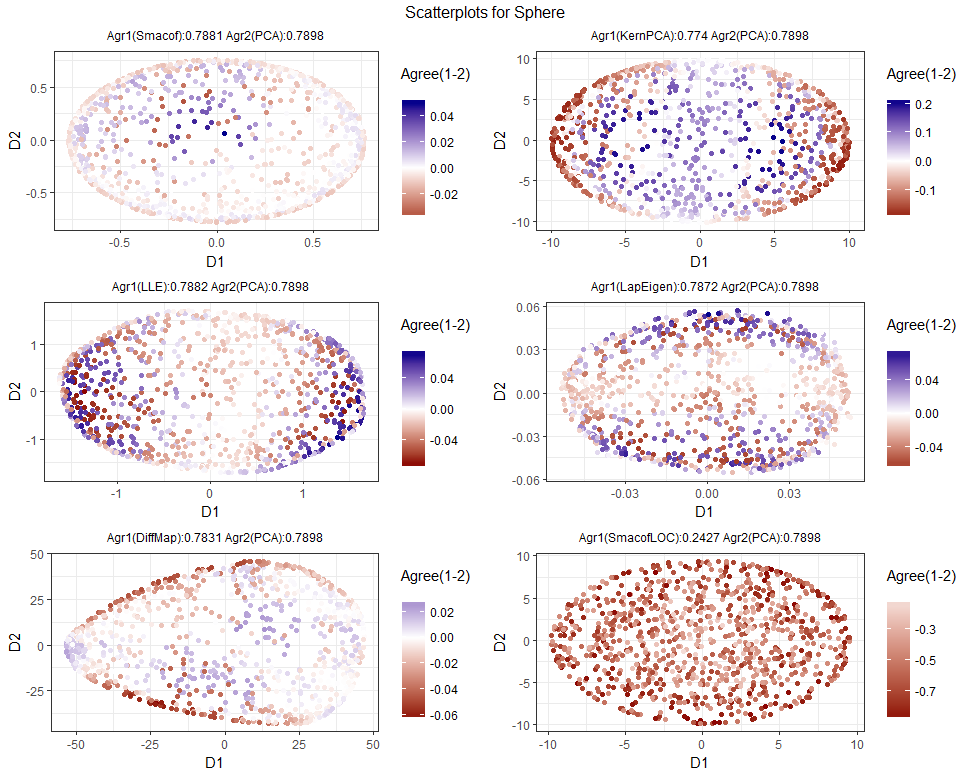}
 \caption{PCA vs. 6 Techniques on Circle for all \textit{k}.}
 \label{fig:2DComp1}
\end{figure}

A similar set of analyses were performed using the 2D scatterplot functions, but this time aimed to compare the basic PCA technique with a range of other techniques on two different manifolds, the sphere and the Swiss roll. The 2D scatterplot function for QVisVis has a similar color scheme to the 3D scatterplot function. The 2D scatterplot allows users to view the shape of the lower dimensional configuration and thus relate the level of agreement to the actual mapping.  The plotted shape is for the first comparison technique. A set of six scatterplots comparing PCA to six other dimensionality reduction techniques for the random circle is given in Figure \ref{fig:2DComp1}.
Across all \textit{k}, PCA performs quite well against the other methods, which is to be expected, as it is a global method.  Smacof and PCA have very similar results, which can be determined from both the similarity in overall agreement rate and the light colors for the individual items, which indicates little difference in agreement. Most techniques render a 2D circle from the sphere, though for the diffusion maps method this was somewhat distorted.  Local MDS performs very badly, giving very poor global recovery, with nearly every item shaded dark red, indicating much poorer item-wise agreement than PCA.

\begin{figure}[htb]
 \centering 
 \includegraphics[width=0.9\columnwidth]{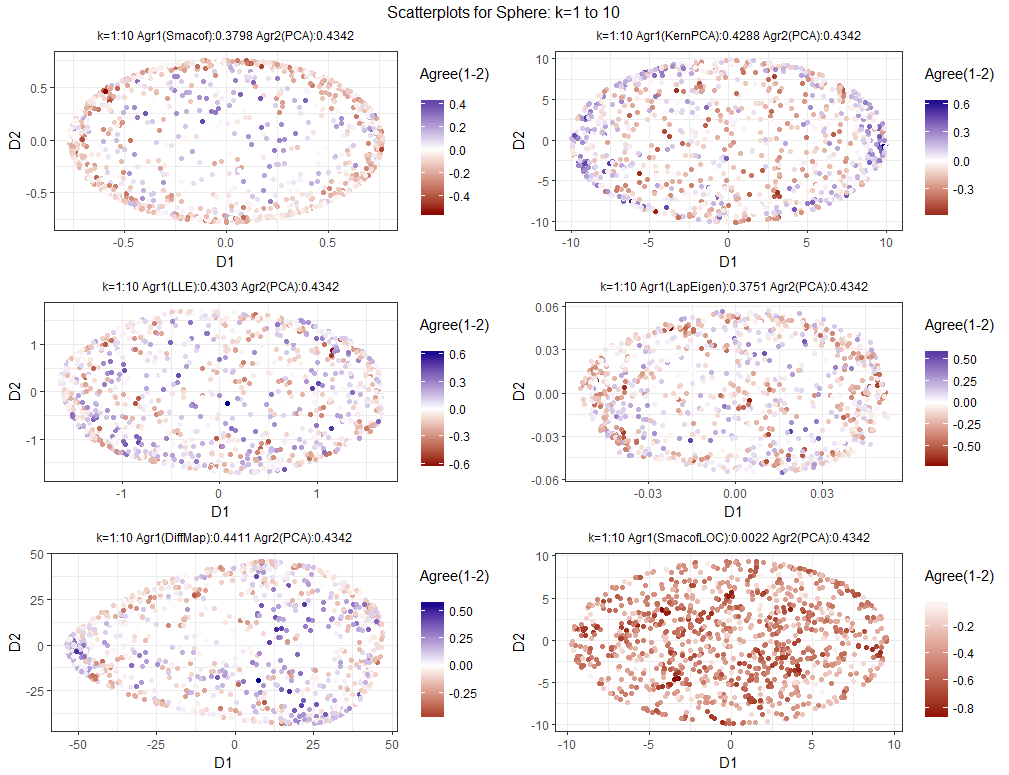}
 \caption{PCA vs. 6 Techniques on Circle for $k=1\cdots10$.}
 \label{fig:2DComp2}
\end{figure}

\begin{figure}[htb]
 \centering 
 \includegraphics[width=0.9\columnwidth]{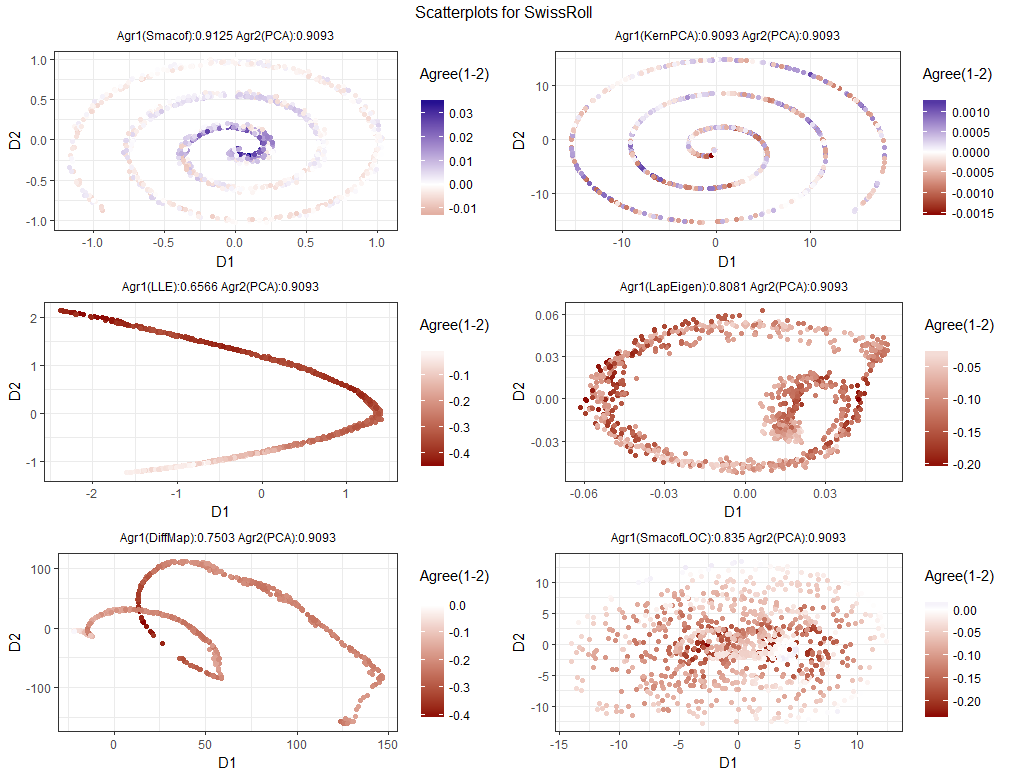}
 \caption{PCA vs. 6 Techniques on Swiss Roll for all \textit{k}.}
 \label{fig:2DComp3}
\end{figure}

As per the last example, the visualization is repeated, but only for $k=1\cdots10$ local neighborhoods. The resulting example is given in Figure \ref{fig:2DComp2}.
The results are a little more even, which is indicated with generally lighter shades of blue and red.  The blues and reds are more evenly distributed, so local neighborhood recovery is less dependent on the area of the visualization than global neighborhood recovery. Again, Local MDS performed badly, indicating that the neighborhood size was perhaps set too low.  The diffusion maps technique has the best overall neighborhood recovery, which demonstrates that a technique can have trouble recovering an overall shape, but can still have good neighborhood recovery.
The process was repeated for the Swiss roll dataset. The all \textit{k} visualization for the Swiss Roll is given in Figure \ref{fig:2DComp3}.  The Swiss roll is quite often used as a classic example of how global dimensionality reduction techniques can have problems with local neighborhood recovery, as the distances between the concentric layers of the roll can distort the results from dimensionality reduction methods.
Here, PCA again has strong global recovery, though Smacof has slightly better recovery than both PCA and kernel PCA.  PCA and kernel PCA are almost identical, showing that the implemented kernel (Gaussian) had very little effect on the overall embedding.  All the other methods had poor overall recovery.  Some evidence of a ``roll'' can be seem for the Laplacian Eigenmap and the diffusion maps methods, but the exactness of the shape is lost.  
\begin{figure}[htb]
 \centering 
 \includegraphics[width=0.9\columnwidth]{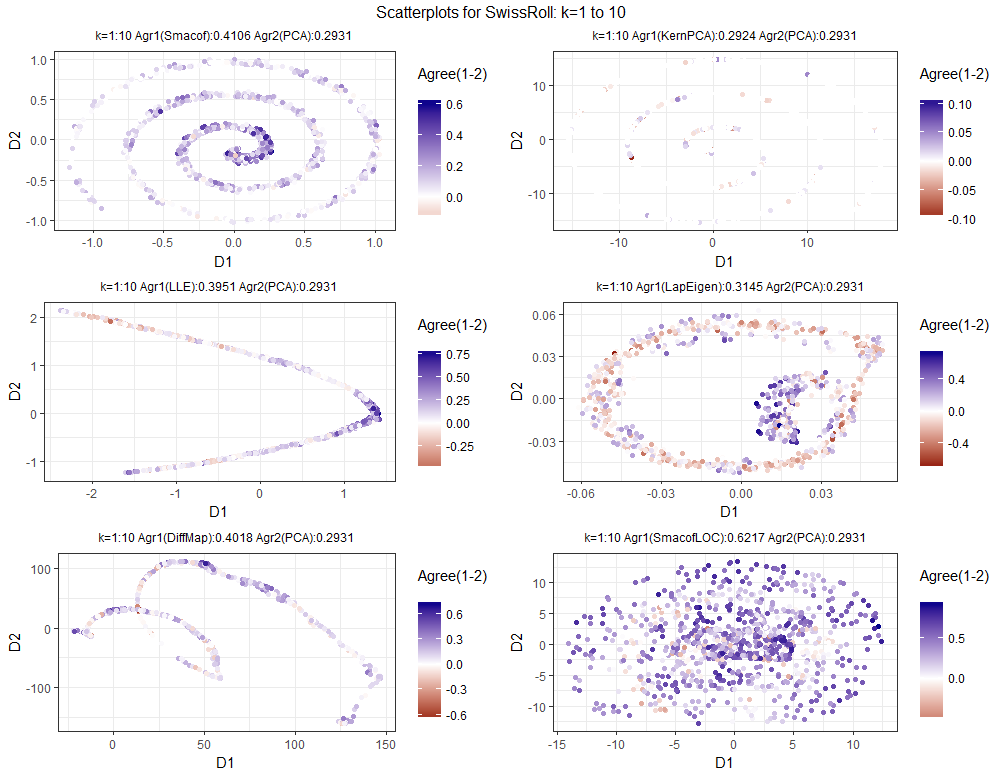}
 \caption{PCA vs. 6 Techniques on Swiss Roll for $k=1\cdots10$.}
 \label{fig:2DComp4}
\end{figure}
For $k=1,\cdots,10$, the seemingly random local MDS solution has by far the best local recovery, while PCA and kernel PCA have quite poor recovery.  The kernel PCA vs. PCA comparison plot has almost disappeared, as the configurations have almost identical item-wise agreement, making most of the points the same color as the background.  The poor performance for PCA may be due to the width dimension of the Swiss roll collapsing onto a two dimensional spiral. This means that items are almost randomly placed on top of one another, giving poor item-wise local recovery.

\subsection{Heatmaps}\label{Heatmaps} 
For the scatterplot figures, the plots varied by comparison technique and (in the first two plots) shape.  It would be possible to create a series of scatterplots that covered different values of \textit{k}, but for 1000 item datasets, multiple values of \textit{k} must be combined in a scatterplot.
Heatmaps have long been used to analyze values over sets of parameters \cite{REF1534} or for comparing quality values across techniques.  For example, \cite{REF1606} give a polygon heatmap for three different solution quality metrics (stress, correlation, and neighborhood preservation), where each technique represents a vertex of the polygon and the points in the interior of the polygon are determined by a weighted sum of the values at the vertices.  Heatmaps have been used to analyze dimensionality reductions solutions, for example, by relating source dimensions to the  lower dimensional projection plane \cite{REF1621}.

In the QVisVis framework, a heatmap can display agreement over both \textit{k} and item dimensions.  Though the ability to see where a point is on the mapping is lost if the points are logically ordered relative to their place on the visualization, some sense of place can be kept.  As per the scatterplots, the items are colored on a blue/red continuum for the agreement of the first technique minus the second technique.

\begin{figure}[htb]
 \centering 
 \includegraphics[width=12cm]{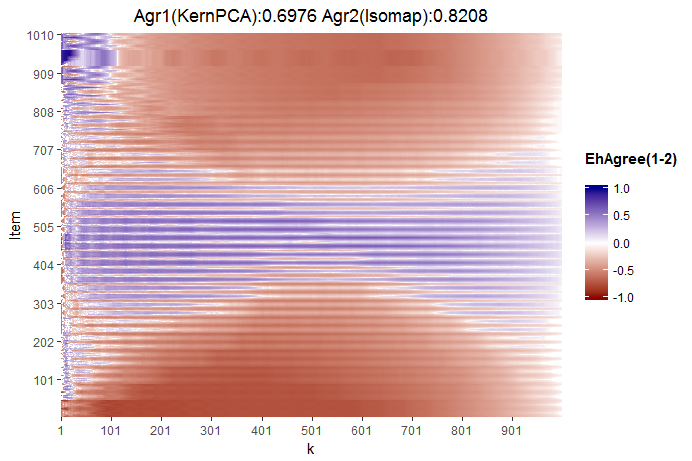}
 \caption{Heatmap for Kernel PCA vs. Isomap on Sphere.}
 \label{fig:Heatmap1}
\end{figure}

The first graph, given in Figure \ref{fig:Heatmap1}, shows a comparative heatmap, showing the results of Kernel PCA vs. Isomap.
\begin{figure}[htb]
 \centering 
 \includegraphics[width=12cm]{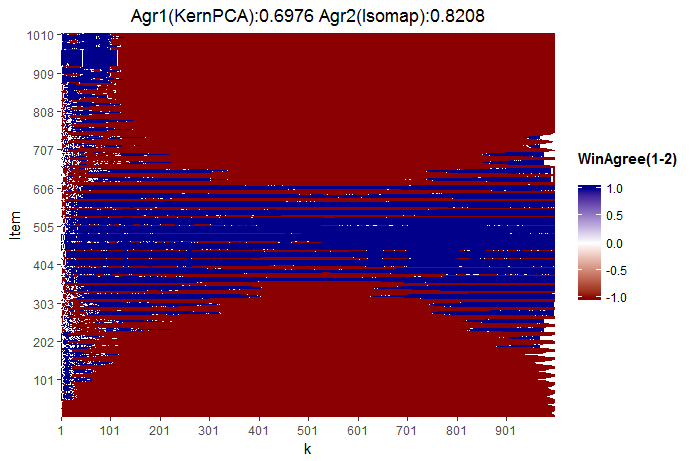}
 \caption{Binary Heatmap for Kernel PCA vs. Isomap on Sphere.}
 \label{fig:Heatmap2}
\end{figure}
The overall agreement across all \textit{k} is 0.8208 for Isomap and 0.6976 for Kernel PCA.  One can see that there is a good deal of consistency across \textit{k}. Most of the items in the center of the configuration have outperformance with Kernel PCA over the whole range of \textit{k}. To give a sharper idea of wins vs. losses, a binary heatmap can be created, where the agreement is coded as 1 if technique 1 wins, -1 if technique 2 wins, and 0 if the agreements are equal. This binary map is given in Figure \ref{fig:Heatmap2}.
\begin{figure}[htb]
 \centering 
 \includegraphics[width=12cm]{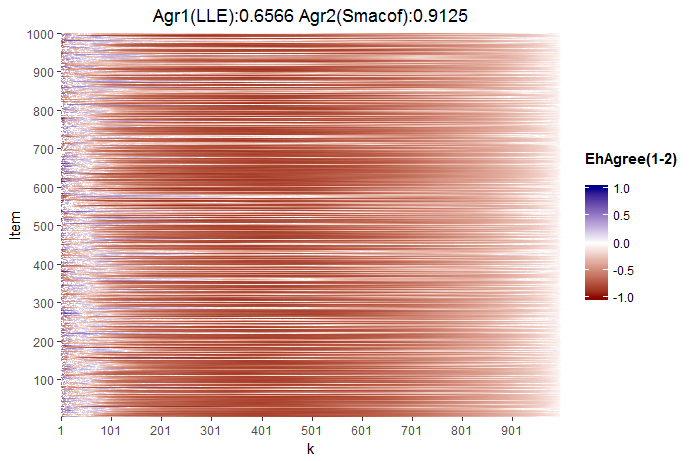}
 \caption{Heatmap for LLE vs. Smacof on Large Diameter Regular Torus.}
 \label{fig:Heatmap3}
\end{figure}
The Heatmap for LLE vs. Smacof on the large diameter regular torus is an interesting one. In terms of overall agreement Smacof dominates LLE quite substantially. However, the techniques are quite evenly matched for low values of \textit{k}. But, as \textit{k} increases, Smacof begins to dominate LLE on all items.  The red colors lessen a little as \textit{k} tends towards $n-1$, but this is to be expected, as for $k=n-1$, agreement values are equal to 1 for all techniques.

\subsection{Heatmaps Overlays}\label{Overlays}
Heatmaps can be overlaid onto the plot of the points.  This has been done to give a gradient map of partial stress \cite{REF1604}.  It is also utilized in the VisCoDeR \cite{REF1602} tool for comparing dimensionality reduction algorithms.  In VisCoDeR, dimensionality projections are animated throughout the process of dimensionality reduction and are points are colored by data cluster.  A greyscale Voronoi diagram that is shaded by local solution quality is displayed behind the projected points. In \cite{REF1622}, a similar Voronoi diagram is utilized, but tears and false neighborhoods in the configuration are given different colors.

\begin{figure}[htb]
 \centering 
 \includegraphics[width=0.9\columnwidth]{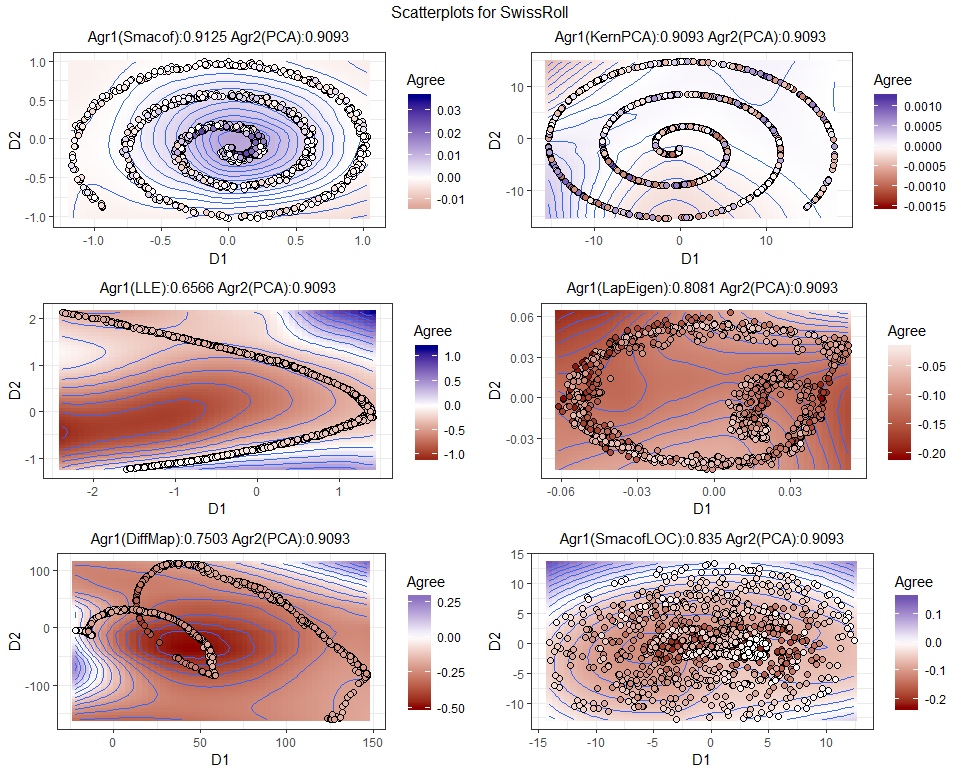}
 \caption{PCA vs. 6 Techniques on Swiss Roll with loess}
 \label{fig:2DCompLoess}
\end{figure}

QVisVis utilizes a local regression and point interpolation (loess) \cite{REF1631,REF1624} approach to overlaying heatmaps onto scatterplots.  This is implemented in the loess() function in R and is used to create a smoothed surface across the space of the scatterplot.  To ensure that the point colors do not blend into the loess colored background, points are given a black outline.  In addition, points can be superimposed onto the smoothed surface plot with external color schemes based on cluster or categorical attribute.  An example is given in Figure \ref{fig:2DCompLoess}.  Here the scatterplot for the Swiss roll dataset given in Figure \ref{fig:2DComp3} is plotted with a loess surface.  The loess surface helps users explore overall patterns in performance.  For example in the Smacof vs. PCA comparison, Smacof has stronger agreement towards the center of the roll, but worse agreement towards the outside of the roll.  
\subsection{Performance Lift Area Plots}\label{LiftPlots}
\begin{figure}[htb]
 \centering 
 \includegraphics[width=0.9\columnwidth]{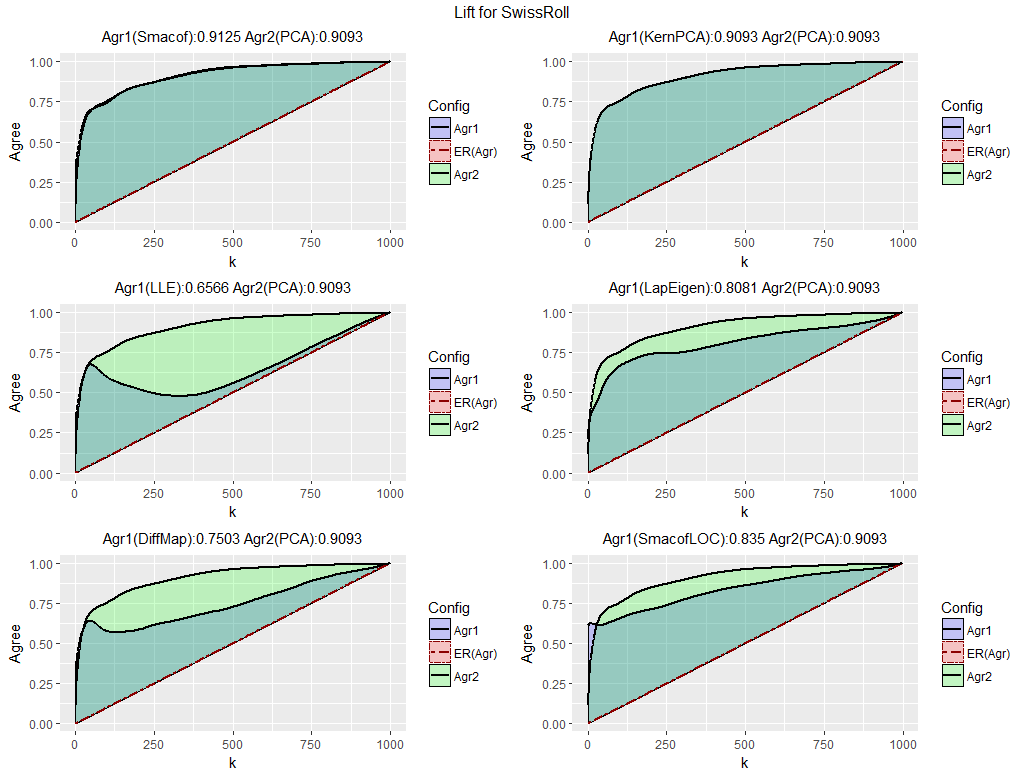}
 \caption{Performance Lift for PCA vs. Other Techniques on Swiss Roll.}
 \label{fig:PerfLift1}
\end{figure}

The performance lift area plots are designed to show the $\psi$ overall agreement metric in terms of area above expected agreement.  The area plots implemented in QVisVis can be utilized for comparing single dimensionality reduction techniques with random agreement or for comparing dimensionality reduction techniques against one another. Performance lift area plots for six different dimensionality reduction techniques vs. the Swiss roll are given in Figure \ref{fig:PerfLift1}.
On each diagram, each of the lift techniques is assigned a color.  The area between a technique's lift curve and expected agreement is colored in that color.  If two techniques overlap then a color intermediate to the two colors is shown. One can see that across all \textit{k}, PCA outperforms most of the techniques, except for Smacof, which maintains a slight, but consistent outperformance. A few of the diagrams are particularly interesting. After $k=100$, LLE seems to almost implode and it's performance rapidly reverts back to almost random performance.  SmacofLOC has a slight area of outperformance for very small \textit{k}, but then is dominated by PCA, which again emphasizes the local nature of the algorithm. 

\section{A Consumer Mapping Example with t-SNE}\label{Discussion}
\begin{figure}[htb]
 \centering 
 \includegraphics[width=0.9\columnwidth]{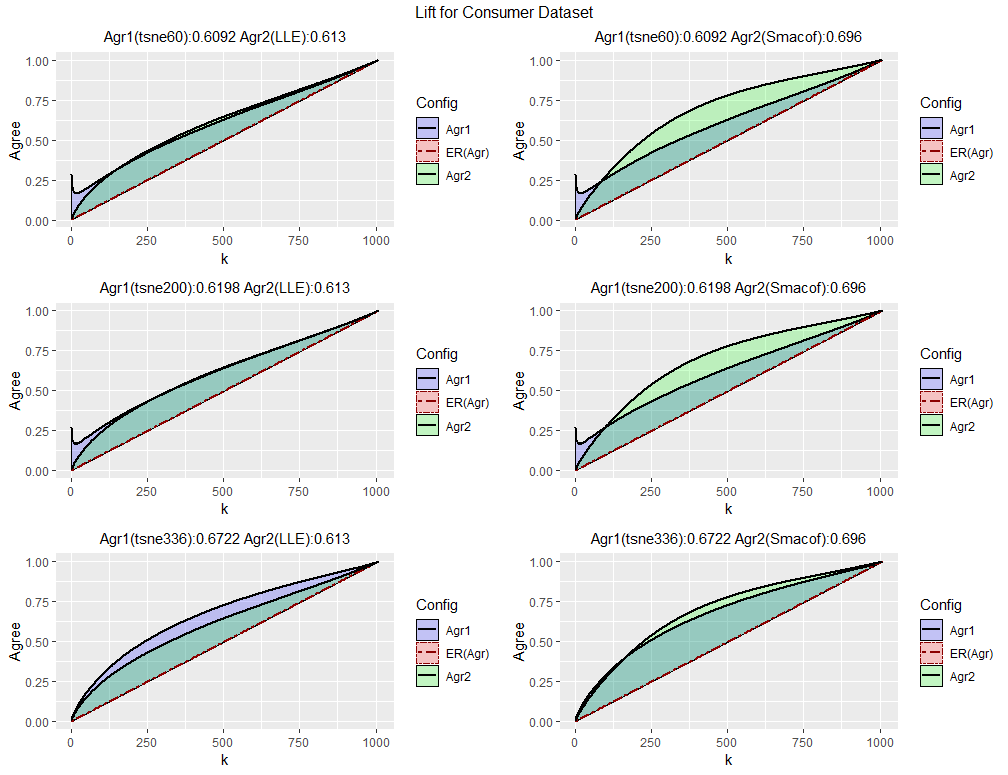}
 \caption{Performance Lift for Consumer Data}
 \label{fig:PerfLiftYP}
\end{figure}

Consumer mapping is a widely applied method to help segment consumers and understand consumer responses to marketing activities \cite{REF152}. A range of multidimensional scaling based methods have been utilized in marketing mapping applications \cite{REF1625,REF1633}.  To demonstrate the use of QVisVis on real-world data, several algorithms were tested on a set of consumer data\footnote{Data can be downloaded from \url{https://www.kaggle.com/miroslavsabo/young-people-survey}}, which gives music, movie, and lifestyle preferences for a young adults.  The dataset has 1010 items and 135 questions, each requiring a Likert scale response from $1\dots5$.  It has a few missing values, which account for 0.4\% of the data.  The t-SNE method \cite{REF1626,REF1632}, which is a variant of stochastic neighborhood embedding \cite{REF1627}, was implemented, as it is optimized to give good local neighborhood recovery.  It has been used in marketing mapping applications, e.g., \cite{REF1249}, where local neighborhood recovery is important.  In this example, the QVisVis framework was used for tuning ``perplexity'' , which is a parameter that is approximately one third of the number of nearest neighbors included when calculating the lower dimensional embedding. Distance MDS implemented with Smacof and LLE were selected as benchmarks, as they are designed for global and local recovery respectively.  The t-SNE method was implemented using the ``Rtsne'' wrapper package.  Initial tuning of the perplexity parameter \textit{p} showed strong values of local neighborhood recovery  ($k=1\dots50$) for  $p=60$.  This value was used, along with the maximum possible value ($p=\lfloor n/3\rfloor=336$) and an intermediate value ($p=200$) in order to examine the trade-off between the perplexity parameter and performance.  Parameter tuning was also performed for LLE by using the ``calc\_k'' function in the ``lle'' R package. To create an overall comparison, performance lift diagrams were created to compare recovery across all k.  For Smacof, basic MDS was implemented with an ordinal distance transformation. The three parameter settings of t-SNE were each compared with each of Smacof and LLE, giving 6 diagrams. The resulting visualizations are shown in Figure \ref{fig:PerfLiftYP}.  

One can see that t-SNE with the smaller perplexity values outperforms both LLE and Smacof when recovering nearest neighbors.  t-SNE dominates LLE across all values of k for $p=200$ and $p=336$, though for larger values of \textit{k}, LLE outperforms t-SNE with perplexity of 60.  t-SNE outperforms Smacof for small values of  \textit{k} (up to around $k=100$), but for large values of \textit{k}, Smacof outperforms t-SNE.  This performance effect is present for all three perplexity values, though it decreases as  \textit{p} increases.  To examine this effect further, a set of heatmaps was produced comparing t-SNE with $p=60$ with Smacof.  Comparisons were created for $k=1\dots50$, $k=51\dots100$, $k=101\dots150$, and $k=151\dots200$ to give two heatmaps each on either side of the performance inversion between t-SNE and Smacof.  As could be expected, as \textit{k} increases, the heatmaps go from mostly blue (t-SNE outperformance) to red (Smacof outperformance). The aggregate aggrement statistics emphasize this pattern. For $k=1\dots50$, t-SNE has over double the agreement of Smacof (0.1823 vs. 0.0885), while for $k=151\dots200$, agreement for t-SNE was substantially lower than for Smacof. The heatmaps are given in Figure \ref{fig:HeatYP}.
\begin{figure}[htb]
 \centering 
 \includegraphics[width=0.9\columnwidth]{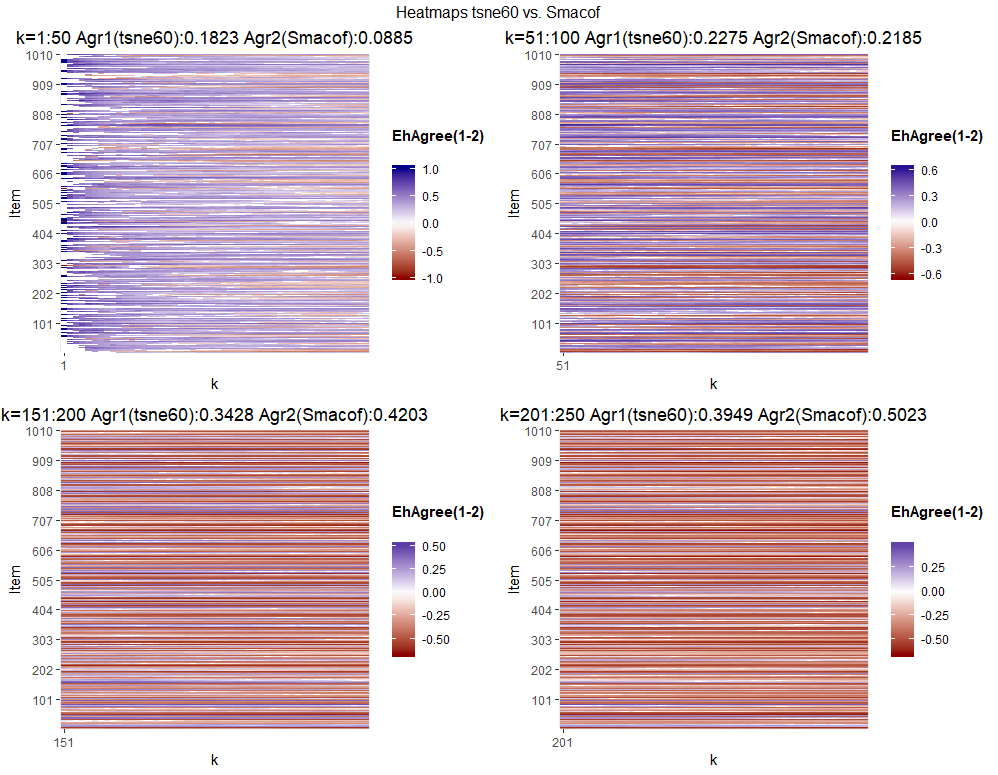}
 \caption{Heatmaps for Consumer Data}
 \label{fig:HeatYP}
\end{figure}
To further look at the differences in performance, scatterplots with loess smoothing were plotted comparing each t-SNE solution with the Smacof solution.  Three plots were created for ``local'' $k=1\dots50$ to show near neighbor recovery and three plots were taken showing ``global'' agreement across all \textit{k}. For the three local plots, the t-SNE solution was used and for the three global plots, the Smacof solution was used.  As expected, t-SNE strongly dominates in the local plots and Smacof outperforms in the global plots.  There also seems to be a slight effect, where Smacof does relatively better towards the center of the configuration and t-SNE does better towards the edge of the configuration.

\begin{figure}[htb]
 \centering 
 \includegraphics[width=0.9\columnwidth]{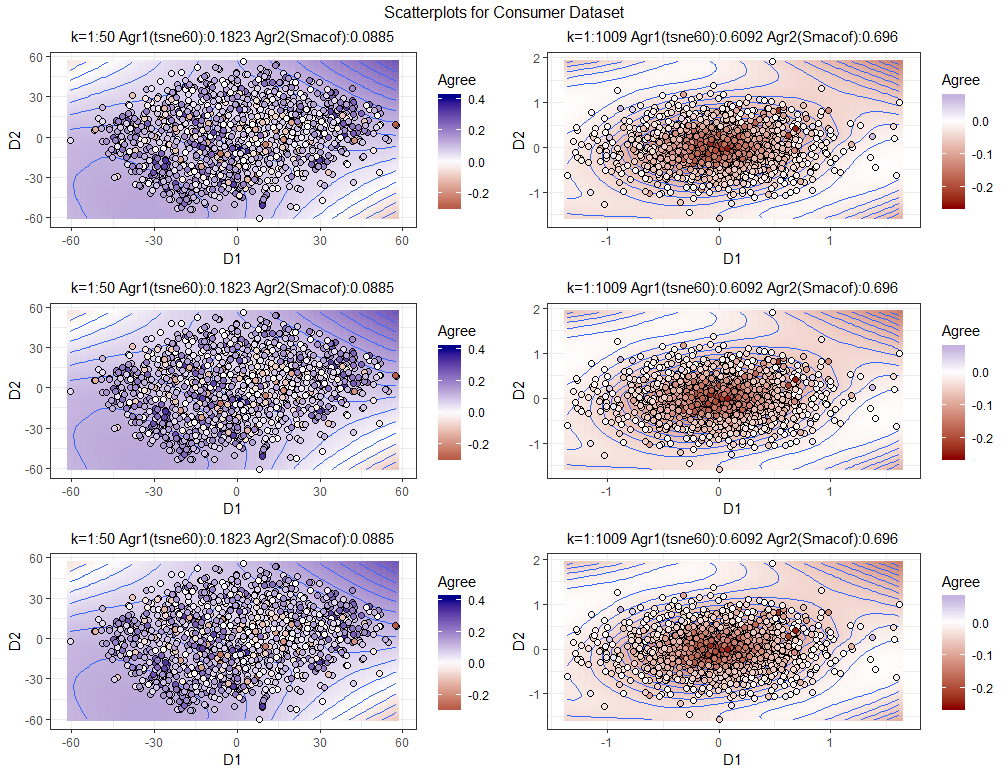}
 \caption{Heatmaps for Loess Plots}
 \label{fig:HeatYP}
\end{figure}

Overall, in a situation where local neighborhood agreement is important, for example, where a marketing manager who wishes to examine relationships betweeen individual customers, then the results show that t-SNE has the best recovery.  However, if global agreement is important, for example, if a marketing manager wishes to examine segments across an entire customer dataset, ordinal MDS implemented with Smacof gives very good performance (though only slightly better than t-SNE with the maximum value of \textit{p}).  In the context of calculating agreement, this example showed how multiple visualizations can give different insights into comparative performance, across both neighborhood size and the geography of the solution configuration. 

\section{Discussion and Future Work}\label{Discussion}
This paper presents a review and synthesis of rank-based methods of analyzing dimensionality reduction performance. This review is used to build a framework for analyzing and visualizing dimensionality reduction performance entitled QVisVis. It is a practical framework, designed to help evaluate dimensionality reduction techniques and to give insight into the performance of the techniques beyond a simple numeric quality metric. The overall framework allows for dimensionality reduction algorithm performance to be analyzed versus random agreement and against other techniques. It allows users to examine the local vs. global performance trade-off and to examine performance in different areas of the source (if 3 dimensional) and output visualizations.

Much further work can be done on the framework and software. The current visualizations show solution agreement, but other relevant information, such as measures of item centrality of the point could be shown \cite{REF1599}.  The software has the ability to draw visualizations across different technique parameter settings, but this ability was not emphasized in this paper and future work could help develop visual based methods of parameter tuning. The visualizations are currently static. Animated visualizations could parsimoniously help users see how visualization performance changes over time and a wide variety of neighborhood sizes and parameter settings. Finally, a wider range of quality measures could be implemented.  For example. the NIEQA method \cite{REF1535} shows how well item neighborhoods are preserved under normalization.  

In this paper, the idea of solution quality is based upon objective, quantitative criteria.  However, in practical visualization contexts, human perception of visualization is important. There have been several article on incorporating human perceptions of visualizations into evaluation of visual embeddings, for example, in \cite{REF1607},  a user survey is used to analyze human perceptions of visualizations with respect to quantitative quality metrics. In a similar fashion, other work has gathered survey data on the quality of cluster seperation for different types of scatterplot \cite{REF1608} and on testing how well visualizations allow users to carry out tasks, such as analyzing the number of clusters and relative distances in visualizations \cite{REF1610}. In \cite{REF1623}, human perceptions of scatterplot similarity were compared with scagnostic measures and correlations were found to be low ($r<0.26$), indicating that much of what humans observe when assessing scatterplots cannot be explained by simple shape based metrics. The combination of quantitative quality metrics and the analysis of human interaction with visualizations and could lead to further insights into visualization quality.  

The scatterplots created in the QVisVis framework can be used to show i) the relative performance of different techniques, ii) performance vs. different parameterizations of techniques, or iii) performance across different subsets of neighborhood recovery.  For parsimony, the examples in this paper are quite simple. However, by combining i, ii, and iii, we could get many hundreds of scatterplots.  These could be displayed in scatterplot matrices (SPLOMs), but this is not practical when there are large number of scatterplots. The method of scagnostics, which was proposed by John Tukey \cite{REF1615}, involves analyzing scatterplots by shape based characteristics such as the minimum spanning tree, convex hull, or ``skeleton'' \cite{REF1612} formed by the scatterplots and by the features exhibited by the scatterplots such as the convexity, monotonicity, and sparsity of the scatterplots \cite{REF1611}. These characteristics can then be used to explore and cluster scatterplots \cite{REF1613} or mathematical transforms of scatterplots \cite{REF1614}.  Combining agreement metrics with overall scatterplot scagnostics could give a deeper understanding of the features and quality of dimensionality reduction algorithms.

\FloatBarrier


\bibliographystyle{spmpsci}      
\bibliography{INFOVIS2018}   

\end{document}